\documentclass[sigconf=false, nonacm=true, review=false, anonymous = false,screen=true]{acmart}
\usepackage{multirow}
\usepackage{afterpage}

\copyrightyear{2023}
\acmYear{2023}
\setcopyright{rightsretained}
\acmConference[GECCO '23 Companion]{Genetic and Evolutionary Computation
Conference Companion}{July 15--19, 2023}{Lisbon, Portugal}
\acmBooktitle{Genetic and Evolutionary Computation Conference Companion
(GECCO '23 Companion), July 15--19, 2023, Lisbon,
Portugal}\acmDOI{10.1145/3583133.3596419}
\acmISBN{979-8-4007-0120-7/23/07}

\usepackage[inline]{newrevisor}
\usepackage{comment}

\usepackage{soul}

\begin{document}
\title[Analysis of modular CMA-ES on strict box constrained problems]{Analysis of modular CMA-ES on strict box-constrained problems in the SBOX-COST benchmarking suite} 

\author{Diederick Vermetten}
\affiliation{%
    \institution{LIACS, Leiden University}
    \streetaddress{Niels Bohrweg 1}
    \city{Leiden}
    \postcode{2333}
    \country{The Netherlands}}
\email{d.l.vermetten@liacs.leidenuniv.nl}
\orcid{0000-0003-3040-7162}

\author{Manuel López-Ibáñez}
\affiliation{%
    \institution{University of Manchester}
    \city{Manchester}
    \country{UK}}
\email{manuel.lopez-ibanez@manchester.ac.uk}
\orcid{0000-0001-9974-1295}

\author{Olaf Mersmann}
\affiliation{%
  \institution{TH Köln}
  \city{Köln}
  \country{Germany}}
\email{olaf.mersmann@th-koeln.de}
\orcid{0000-0002-7720-4939}

\author{Richard Allmendinger}
 \affiliation{%
    \institution{University of Manchester}
    \city{Manchester}
    \country{UK}}
\email{richard.allmendinger@manchester.ac.uk}
\orcid{0000-0003-1236-3143}

\author{Anna V. Kononova}
\affiliation{%
      \institution{LIACS, Leiden University}
      \streetaddress{Niels Bohrweg 1}
      \city{Leiden}
      \postcode{2333}
      \country{The Netherlands}}
\email{a.kononova@liacs.leidenuniv.nl}
\orcid{0000-0002-4138-7024}
\renewcommand{\shortauthors}{Vermetten, et al.}

\begin{abstract}
  Box-constraints limit the domain of decision variables and are common in real-world optimization problems, for example, due to physical, natural or spatial limitations. Consequently, solutions violating a box-constraint may not be evaluable. This assumption is often ignored in the literature, e.g., existing benchmark suites, such as COCO/BBOB, allow the optimizer to evaluate infeasible solutions. This paper presents an initial study on the strict-box-constrained  benchmarking suite (SBOX-COST), which is a variant of the well-known BBOB benchmark suite that enforces box-constraints by returning an invalid evaluation value for infeasible solutions. Specifically, we want to understand the performance difference between BBOB and SBOX-COST as a function of two initialization methods and six constraint-handling strategies all tested with modular CMA-ES. We find that, contrary to what may be expected, handling box-constraints by saturation is not always better than not handling them at all. However, across all BBOB functions, saturation is better than not handling, and the difference increases with the number of dimensions. Strictly enforcing box-constraints also has a clear negative effect on the performance of classical CMA-ES (with uniform random initialization and no constraint handling), especially as problem dimensionality increases. 
\end{abstract}



\maketitle

\section{Introduction}
Box-constraints impose limits on the domain of decision variables and are perhaps the most typical type of constraint in black-box continuous optimization. In real-world problems, the range of decision variables is often limited by physical, design, resource or policy bounds that are known to the decision maker a priori. Solutions outside those bounds, i.e., violating the box-constraints, are not only unacceptable, but it is impossible to evaluate the objective function at these points. Unfortunately, many optimization algorithms ignore this assumption and evaluate solutions violating box-constraints. Moreover, benchmarking suites used for comparing algorithms, such as COCO/BBOB~\cite{hansen2020coco}, return the actual objective value for such solutions, thus helping algorithms that violate box-constraints inform their search. Consequently, the algorithm insights gained on such benchmarking suites may not hold when faced with a problem where box-constraints are enforced. 

Here, we consider a benchmark SBOX-COST that enforces box-constraints by returning the same invalid value ($\infty$) for any infeasible solution, thus the algorithm cannot use infeasible solutions to inform the search. In practice, constraints that return an invalid or even no value when violated can be found, for example, in expensive optimisation where ephemeral resource constraints (ERCs)~\cite{allmendinger2013handling} define availabilities of resources needed to carry out real-world experiments (e.g. of physical, biological or chemical nature, or even computational experiments requiring licenses, software or other computational resources) hence a (potentially feasible) solution violating an ERC cannot be evaluated resulting no objective function value; another example are safety constraints, which are defined in the objective space and set (known or unknown) lower bounds (assuming minimisation) on the objective function values~\cite{kim2021safe} encapsulating scenarios where the evaluation of a very poor solution (policy or strategy) causes an irrecoverable loss (e.g., breakage of a machine or equipment, or life threat) and hence potentially no objective function value. 

We evaluate the effect that strict-box-constraints have on the performance of some variants of the Covariance Matrix Adaptation Evolution Strategy (CMA-ES). CMA-ES~\cite{Hansen.1996} is a very popular heuristic optimisation algorithm for continuous optimisation problems. CMA-ES is considered state-of-the-art in evolutionary computation and has been adopted as one of the standard tools for continuous optimisation in many research labs. There are many variants of CMA-ES developed through the years and different implementations of sub-components such as the sampling strategy and the boundary correction method~\cite{vermetten_gecco2022}.

In recent works, the different modules and configurations of CMA-ES are explored and analysed based on their performance~\cite{de2021tuning}. In that research, a modular CMA-ES framework is presented, representing a plethora of different CMA-ES configurations. In this paper, the modular CMA-ES framework is used to analyze the effect of enforcing box-constraints in standard BBOB problems, as done in the newly proposed SBOX-COST benchmark. 

The next section describes the methodology followed in this study. Results are discussed in Section~\ref{results}, and conclusions are drawn in Section~\ref{conclusion}.

\section{Methodology}
This section outlines the methodology adopted in this study including the SBOX-COST benchmarking suite and the CMA-ES algorithm considered in the experimental study. 

\subsection{SBOX-COST benchmarking suite}

The Strict Box-Constraint Optimization Studies benchmark\footnote{\url{https://github.com/sbox-cost}} (SBOX-COST) is benchmark suite for optimisation heuristics that aims to  more closely represent real-world problems with box-constraints.  SBOX-COST is a version of the well-known continuous BBOB benchmarking suite~\cite{hansen2020coco} with the same functions subjected to two modifications: 
\begin{enumerate}
    \item enforced \textit{strict box-constraints}, i.e., points evaluated outside of the functions' domain $[-5,5]^d$ are considered infeasible and evaluated to $\infty$, providing no guidance to the optimisation process beyond the domain boundaries\footnote{We opt to use $\infty$ instead of \texttt{NaN} to still enable comparison with other solutions.} (where $d$ is search space dimensionality); 
    \item altered \textit{distribution of optima locations} across instances to ensure a more realistic setting: in BBOB most functions have optima located uniformly in $[-4,4]^d$, leaving an unrealistically wide outside perimeter of the functions' domain free of optima on all possible instances. Therefore in SBOX-COST: 
    \begin{itemize}
        \item optima of all functions except those mentioned below are ensured to have uniform distribution across instances within the full domain $[-5,5]^d$ because as $d$ grows, the fraction of the search space located in $[-5, 5]^d \ [-4, 4]^d$ (i.e. the outer perimeter) grows.
        For example, for $d=10$, almost 90\% of the search space is in the perimeter.
        Therefore the original BBOB setup favours algorithms that focus on the center of the search space;
        \item optima of F5 are unchanged from BBOB and are found in one of the corners of $[-5,5]^d$;
        \item distributions of optima of F4, F8, F9, F19, F20, F24 are also left unchanged compared to BBOB -- however, it is worth mentioning that such distributions have been found~\cite{bib:Long2023} not to follow uniform distribution within $[-4,4]^d$ on BBOB and, therefore, on SBOX-COST as well.
    \end{itemize}
\end{enumerate}

\subsection{Modular CMA-ES}\label{sect:cma-es}
To investigate several commonly and less commonly used configurations and variants of the CMA-ES algorithm~\cite{Hansen.1996}, we use the Modular CMA-ES framework~\cite{van2016evolving,de2021tuning}. The framework is open-source and available\footnote{\url{https://github.com/IOHprofiler/ModularCMAES}} as part of the \texttt{IOHprofiler} \cite{doerr2018iohprofiler} environment. 

The modular CMA-ES framework is currently made up of $11$ modules, each having a number of implemented options (in brackets): Active update (2), Elitism (2), Orthogonal Sampling (2), Sequential Selection (2), Threshold Convergence (2), Step-Size Adaptation (7), Mirrored Sampling (3), Quasi-Gaussian Sampling (3), Recombination Weights (2), Restart Strategy (3) and Boundary Correction (6). A detailed description of all available CMA-ES modules and their settings can be found in~\cite{de2021tuning}.

In this paper, to investigate the effect of the introduction of strict box-constraints on the BBOB problems in the SBOX-COST benchmarking suite, we investigate the behaviour of several variants within modular CMA-ES varying: 
\begin{itemize}
    \item \textit{initialization} methods for the center of mass: 
   \begin{itemize}
        \item[$\circ$] \texttt{center}: origin of the space
        \item[$\circ$] \texttt{random}: uniformly at random in the domain
    \end{itemize}
    \item \textit{strategies of dealing with infeasible solutions\footnote{also referred to as a boundary constrained handling method (BCHM)} (SDIS)}~\cite{Caraffini2019,Kononova2020PPSN,bib:DEAnalysis2022}:
    \begin{itemize}
        \item[$\circ$] \texttt{None}: infeasible solutions are allowed within the population   
        \item[$\circ$] \texttt{saturate}~\cite{Caraffini2019}: infeasible coordinate is placed to the closest corresponding bound 
        \item[$\circ$] \texttt{COTN}: replace infeasible solution coordinates with a value resampled from the truncated normal distribution centered on the closest bound
        \item[$\circ$] \texttt{mirror}: replace infeasible solution with inward reflection off the closest domain bound 
        \item[$\circ$] \texttt{toroidal}: replace infeasible solution with inward reflection off the furthers domain bound
        \item[$\circ$] \texttt{unif\_resample}: replace infeasible solution coordinates with a uniformly distributed value within domain bounds.
    \end{itemize}
\end{itemize} 

All \textit{other modules} and parameter settings in modular CMA-ES are set to their defaults specified in~\cite{de2021tuning}. 

The experimental study will investigate three aspects: (i)~performance difference between BBOB and SBOX-COST when initialization methods are combined with two basic constraint-handling strategies, \texttt{None} or \texttt{saturate} (Section~\ref{firstEx}), (ii)~impact on the level of infeasibility from initial sampling  (Section~\ref{initial_sampling}), and (iii)~impact of the other constraint-handling strategies on the stepsize and the number of infeasible solutions (Section~\ref{secondEx}).

\subsection{Experimental setup}
We evaluate algorithms on the 24 functions of both BBOB and SBOX-COST benchmark suites. 
We use an identical setup for both suites: 15 instances per function, 1 run per instance, dimensionality of $d\in\{5, 20\}$, and fitness evaluation budget of $10\,000\times d$. 

Experiments reported in this study are carried out in the \texttt{IOHex\-pe\-ri\-men\-ter} environment~\cite{de2021iohexperimenter}, which implements both benchmarking suites. The data from the described experiments is visualized using \texttt{IOHana\-ly\-zer}~\cite{IOHanalyzer}.

\begin{figure*}[!t]
    \centering
    \includegraphics[height=0.258\textwidth,trim=9mm 21mm 5mm 5mm,clip]{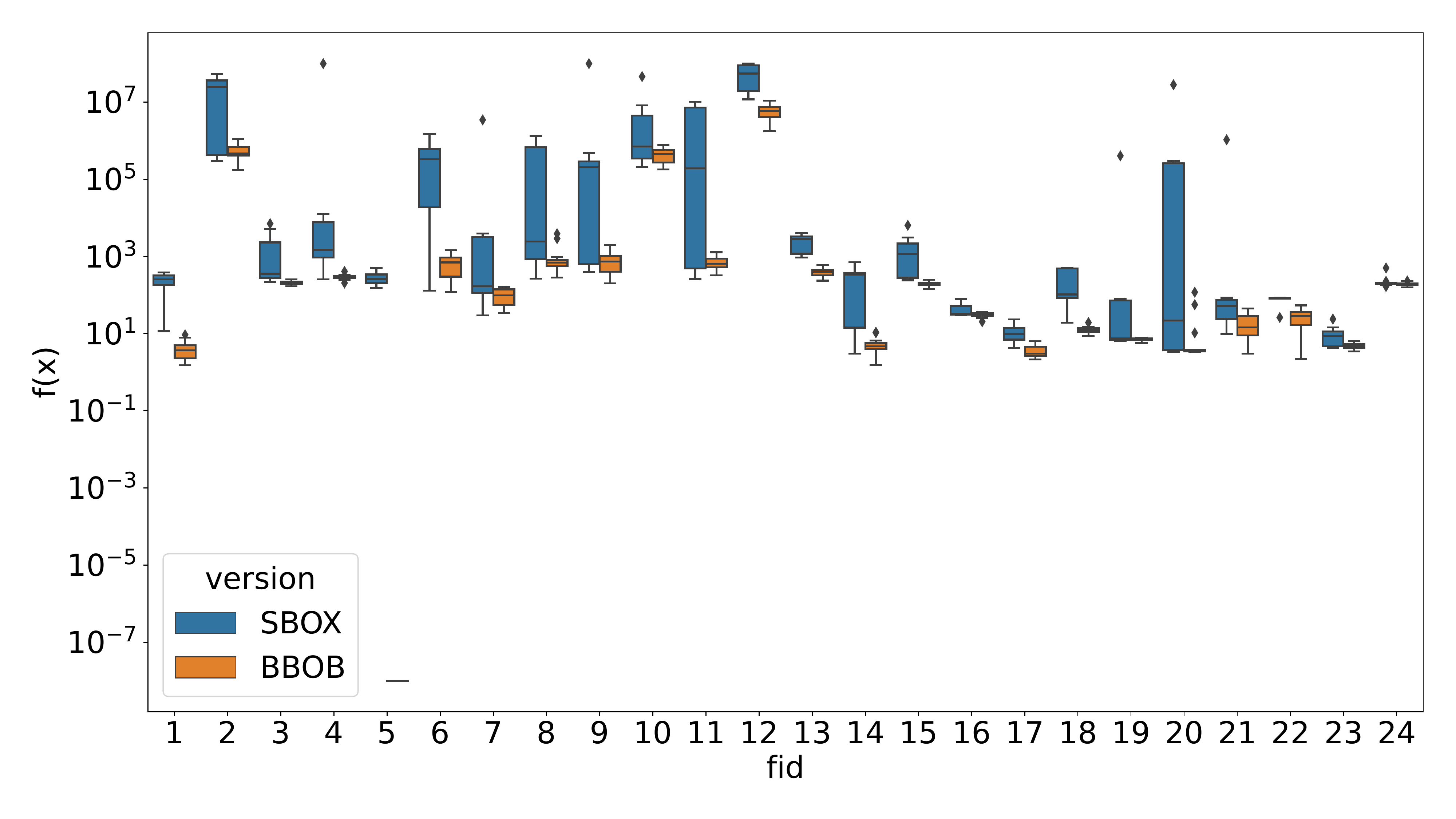}
    \includegraphics[height=0.258\textwidth,trim=17mm 21mm 5mm 5mm,clip]{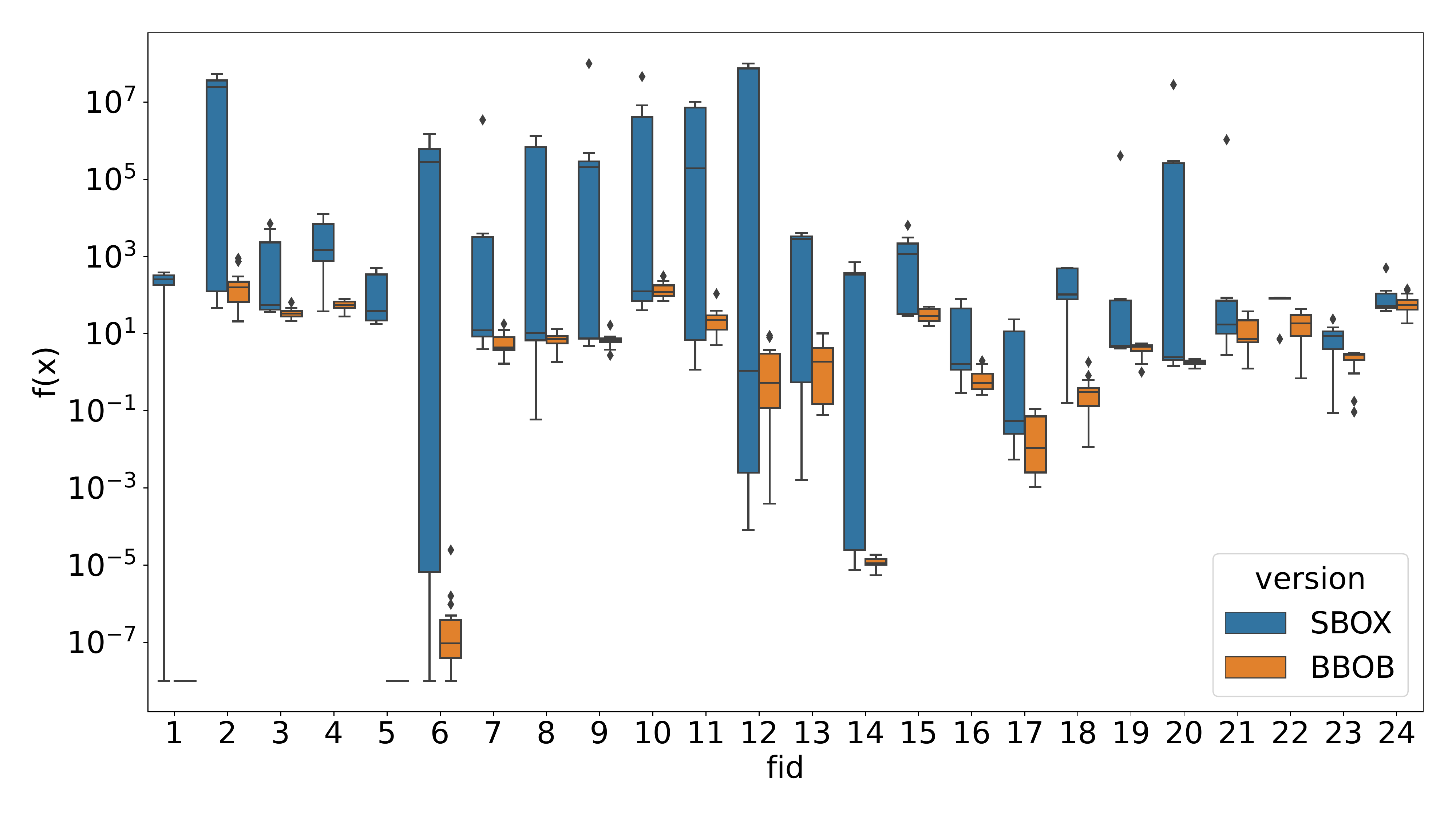}
    \caption{Best-so-far fitness values attained by CMA-ES with uniform initialization and strategy None within 500 (left) and 10000 (right) fitness evaluations on all functions of both suites in 20D.} \label{fig:perf_none}
\end{figure*}

\begin{figure}[!t]
    \centering
    \includegraphics[width=0.9\linewidth,trim=9mm 40mm 13mm 10mm,clip]{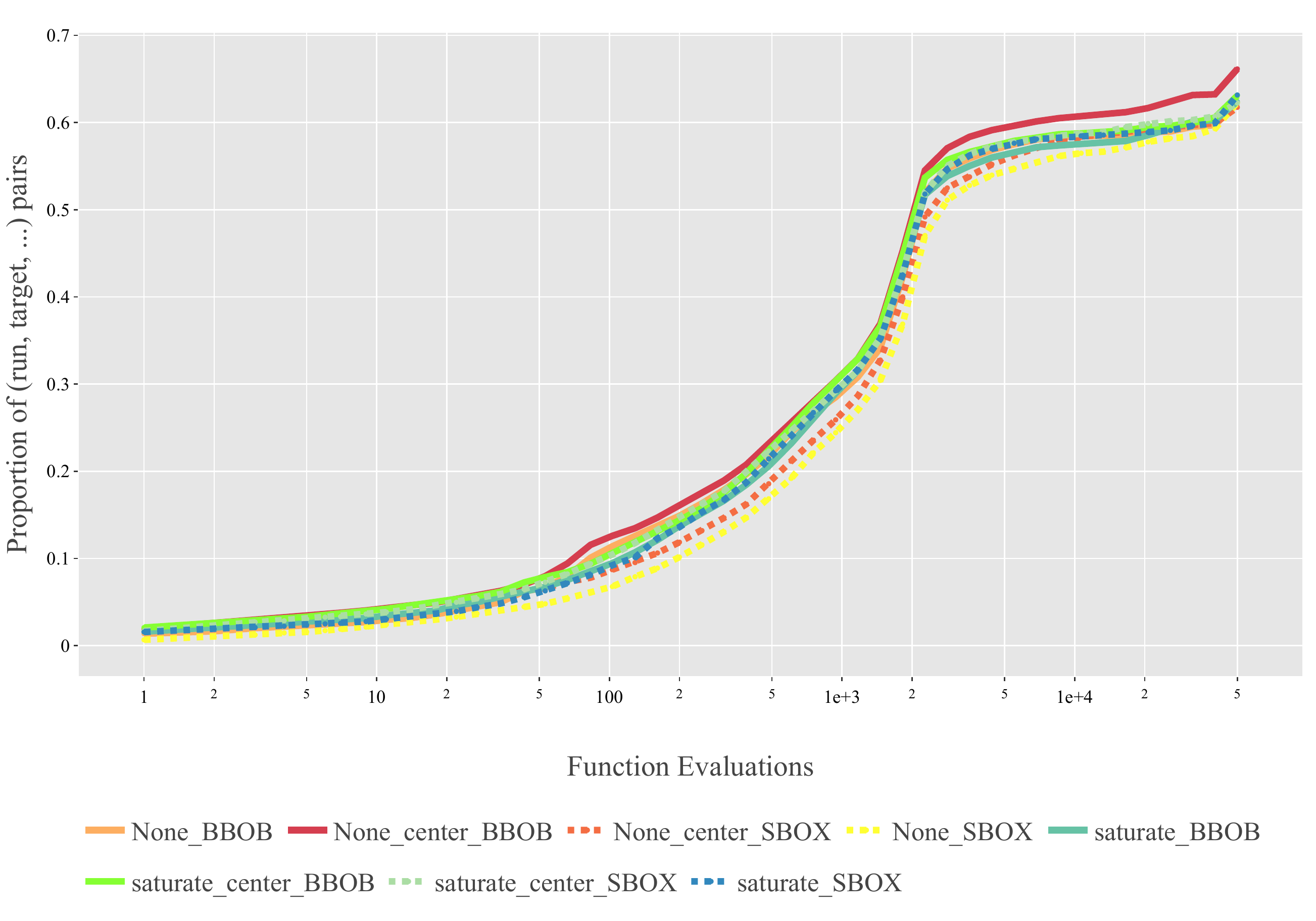}\\
    \includegraphics[width=0.9\linewidth,trim=9mm 40mm 13mm 10mm,clip]{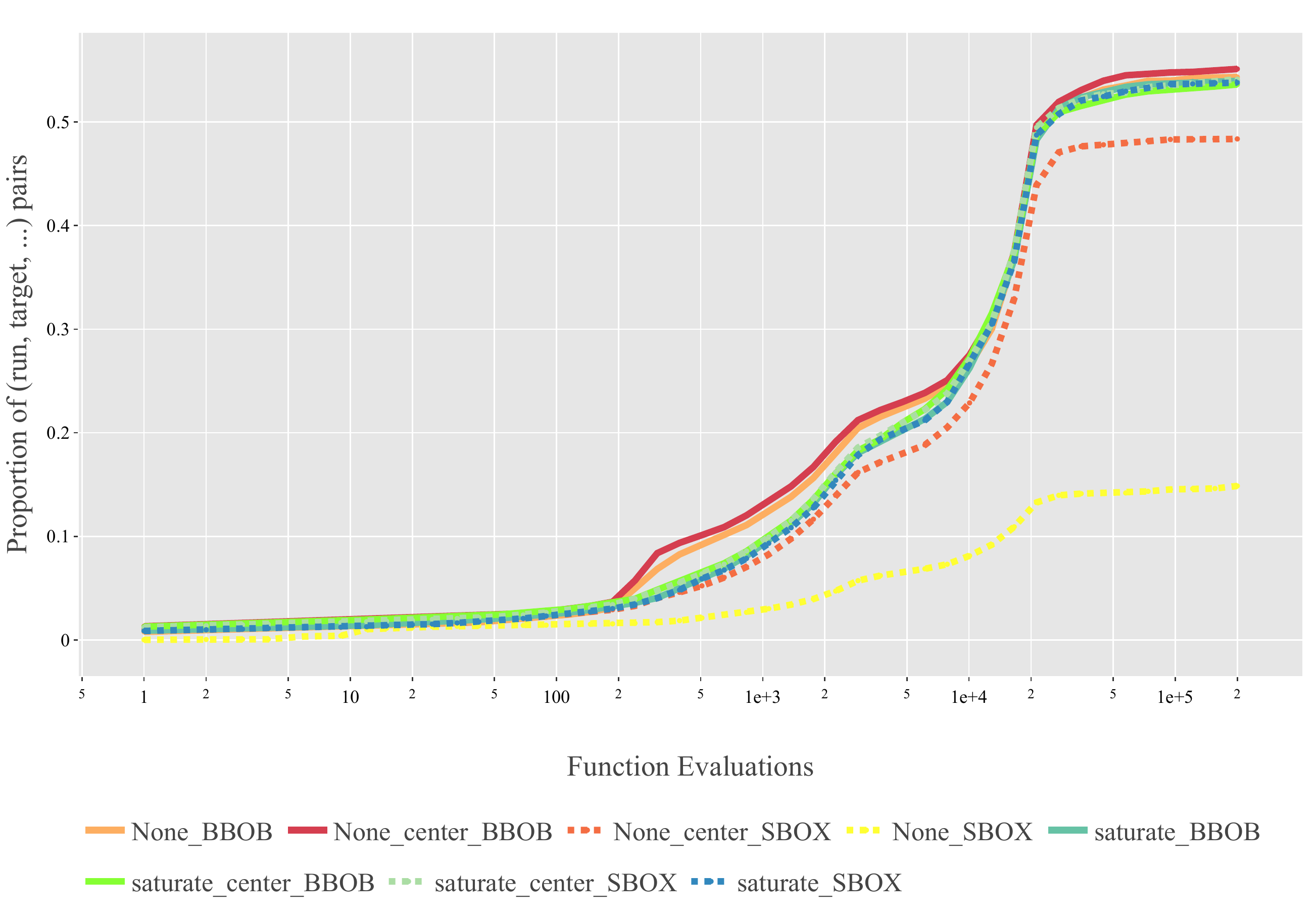}\\
    \includegraphics[width=0.9\linewidth,trim=9mm 6mm 10mm 160mm,clip]{Figures/RT_ECDF_MULT_20D.pdf}
 \caption{Empirical Cumulative Distribution Function for all variants within modular CMA-ES considered in the 1$^{\text{st}}$ experiment, aggregated over the 24 functions of BBOB and SBOX-COST in $d=5$ (top) and $d=20$ (bottom), where the proportion of (run, target) pairs shown on the vertical axis is computed based on targets of 51 log-spaces values between $10^2$ and $10^{-8}$ ('bbob' default in \texttt{IOHanalyzer}). The horizontal axis shows the number of fitness evaluations. The names are a concatenation of the SDIS, initialization mechanism and benchmark suite. 
 }\label{fig:res_ecdf} 
\end{figure}

\begin{figure*}
    \centering
    \includegraphics[width=0.85\linewidth,trim=8mm 70mm 4mm 6mm,clip]{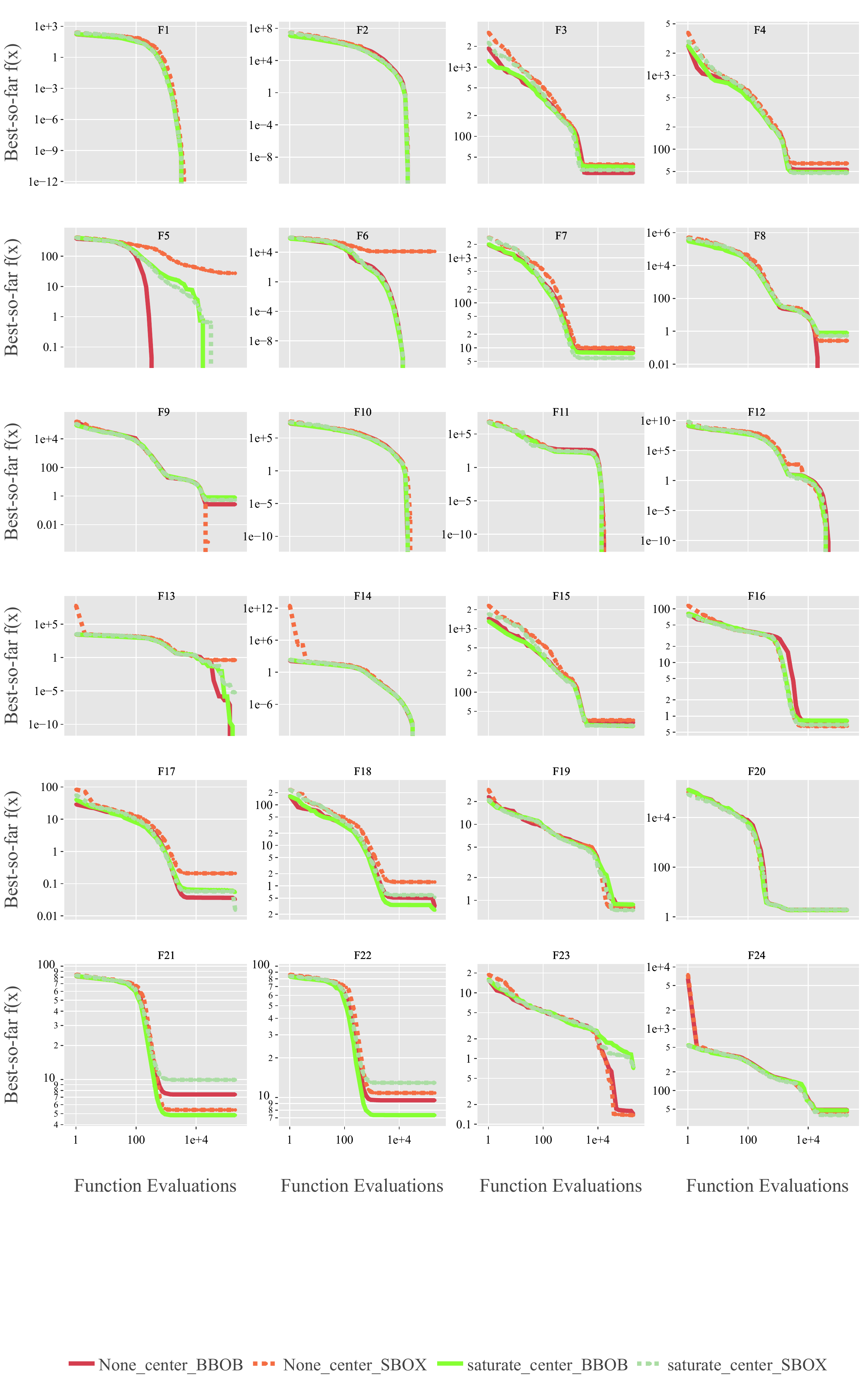}
    \includegraphics[width=0.75\linewidth,trim=8mm 5mm 4mm 410mm,clip]{Figures/FCE_Mult-2023-04-18.pdf}
    \caption{Mean function value over time (fitness evaluations) for the CMA-ES methods with \texttt{center} initialization, for 24 functions in BBOB and SBOX-COST in 20D. The names are a concatenation of the SDIS, initialization mechanism and benchmark suite.
    }\label{fig:res_ert}
\end{figure*}

\section{Results}\label{results}
This section presents and analyses the results of the effect of SBOX-COST (in relation to BBOB) on different initialization and constraint-handling strategies augmented on modular CMA-ES.

\subsection{Performance differences between BBOB and SBOX-COST}\label{firstEx}
We aggregate the performance of 4 selected CMA-ES variants (varying initialization between uniform and center, and varying SDIS between saturation and no SDIS) across all 5-dimensional ($d=5$) problems per suite (BBOB or SBOX-COST) by means of the Empirical Cumulative Distribution function (ECDF). The ECDF computes the fraction of objective value targets that are hit across all runs of each CMA-ES variant up to a given number of function evaluations. We use the BBOB default of 51 logarithmically-spaced targets $\{10^{-8},\dots,10^{2}\}$.

The ECDF for 5D problems is shown in the top plot of Figure~\ref{fig:res_ecdf}. We observe that, while there is almost no difference between the CMA-ES variants on the BBOB suite, the CMA-ES variants without boundary correction perform slightly worse on the SBOX-COST suite than the other variants. The bottom plot of Figure~\ref{fig:res_ecdf} shows the ECDF curves on the 20-dimensional problems (20D, $d=20$). Here, the poor performance of the CMA-ES with random initialization without boundary correction on SBOX-COST is much more obvious.
The explanation is that, in higher dimensions, a Gaussian distribution around a random point will be much more likely to generate points outside the domain, leading to many wasted function evaluations due to the lack of boundary correction, and potentially a disrupted search process. When only a small set of points in the population are infeasible, the rank-based update mechanism of CMA-ES will be relatively unimpacted, but when this fraction grows larger the inability to distinguish between the poor solutions can have a real impact on the algorithm behaviour. 
While initializing CMA-ES in the center of the domain alleviates this problem somewhat, it still performs worse than the variant using the saturate strategy. 
Crucially, this difference cannot be observed in the BBOB suite because box-constraints are not enforced. 
This highlights the need for dealing with infeasible solutions.
Unsurprisingly, similar conclusions have been made regarding other algorithms, such as variants of Differential Evolution~\cite{Caraffini2019,Kononova2022importance}.

To get a more detailed view of the performance on individual functions, we plot the convergence trajectories (mean function value over time) in Figure~\ref{fig:res_ert}. To ease readability, we only show the CMA-ES variants that are initialized in the center of the domain. We can see in Figure~\ref{fig:res_ert} that, for several functions, the performance differences between the algorithms run on SBOX-COST and on BBOB are rather small. In particular, this is the case for functions F4, F19, F20 and F24,
 where the instance generation procedure was not modified from the original BBOB suite. However, the differences on F5 clearly show the impact of the strict box-constraint, with the CMA-ES on BBOB easily solving the problem (achieved by moving outside the domain).

When comparing the performance of CMA-ES with uniform initialization and no boundary handling on the BBOB and SBOX suites (Figure~\ref{fig:perf_none}), we observe that the SBOX suite is more difficult in general and leads to more variance in performance, and this difference increases with higher number of evaluations.

Furthermore, a closer inspection of more detailed summary statistics of runs on BBOB and SBOX-COST (omitted due to space limitations) confirms the observations from Figure~\ref{fig:res_ert}: for many problems, the CMA-ES can overcome the addition of strict box-constraints and changed function initialization, although there are some problems where the changes lead to a noticeable deterioration in performance.

\subsection{Infeasibility from initial sampling}\label{initial_sampling}

\begin{figure*}[!t]
    \centering
    \includegraphics[width=0.65\textwidth,trim=50mm 196mm 25mm 23mm,clip]{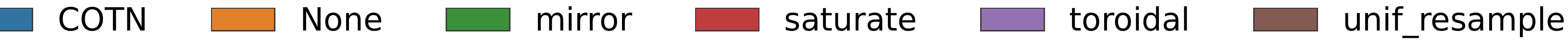}\\
    \includegraphics[height=0.235\textwidth,trim=20mm 21mm 23mm 40mm,clip]{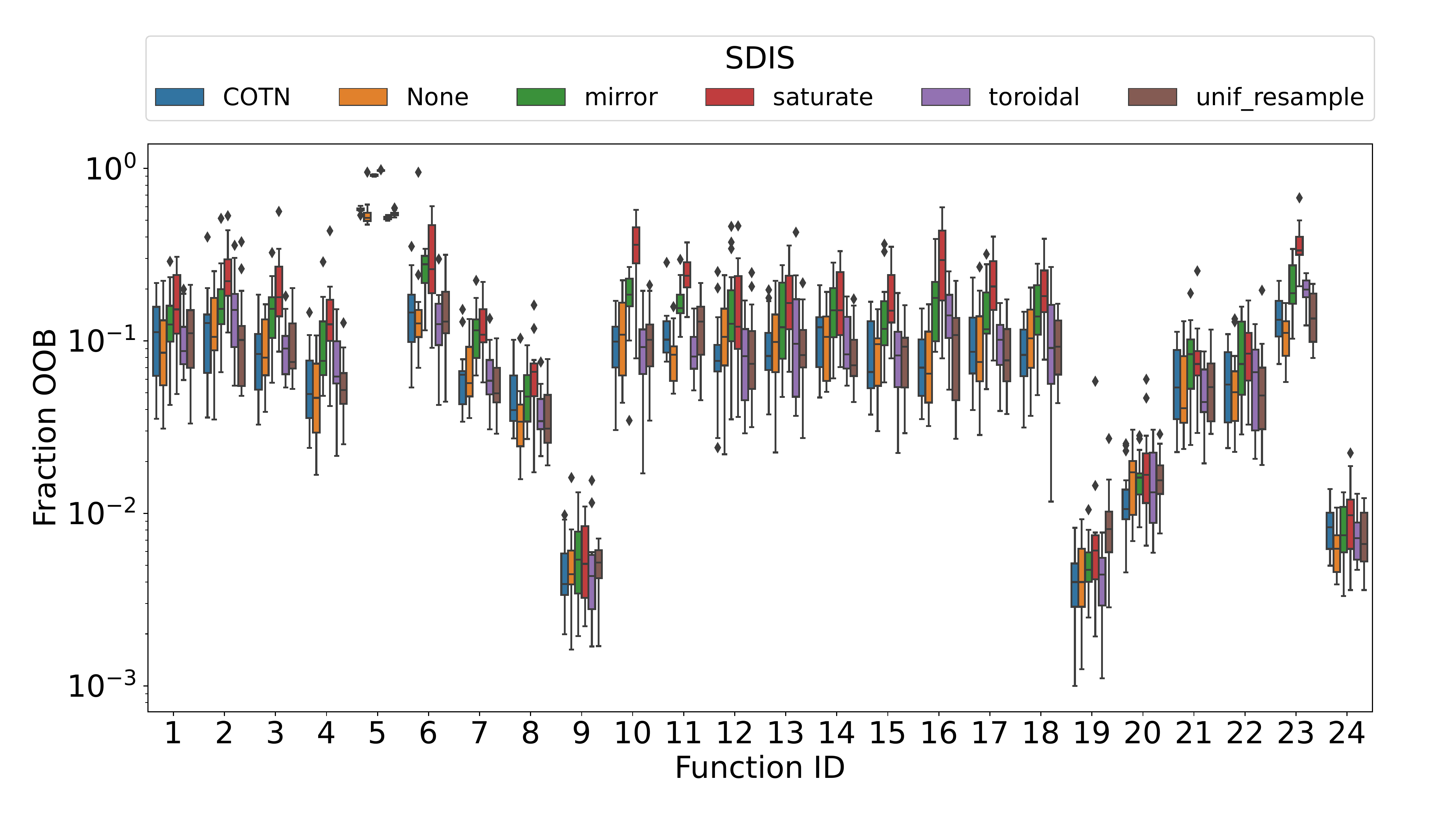}
    \includegraphics[height=0.235\textwidth,trim=38mm 21mm 23mm 40mm,clip]{Figures/OOB_per_SDIS_20D_SBOX.pdf}
    \caption{Ratio of solutions generated outside box-constraints in modular CMA-ES with center (left) or uniform (right) initialization on SBOX-COST problems in 20D for different bound correction strategies.} \label{fig:oob} 
\end{figure*}

As noted in Section~\ref{firstEx}, performance comparisons seem to suggest that CMA-ES with random initialisation of the center of mass (see Section~\ref{sect:cma-es}) generates higher ratios of infeasible solutions. In order to confirm this hypothesis, we plot the ratio of solutions generated outside the bounds of the SBOX-COST problems in Figure~\ref{fig:oob}. Here, we see that the ratio is similar for most problems, of an order of magnitude around $10^{-1}$. An interesting pattern we observe from Figure~\ref{fig:oob} is that the distribution of these ratios for the uniform-initialized CMA-ES with SDIS 'None' is extremely wide, with peaks close to $1$, meaning that almost all evaluated points were unfeasible. 

\begin{figure}[!tb]
    \centering
    \includegraphics[width=0.45\textwidth,trim=25mm 10mm 40mm 27mm,clip]{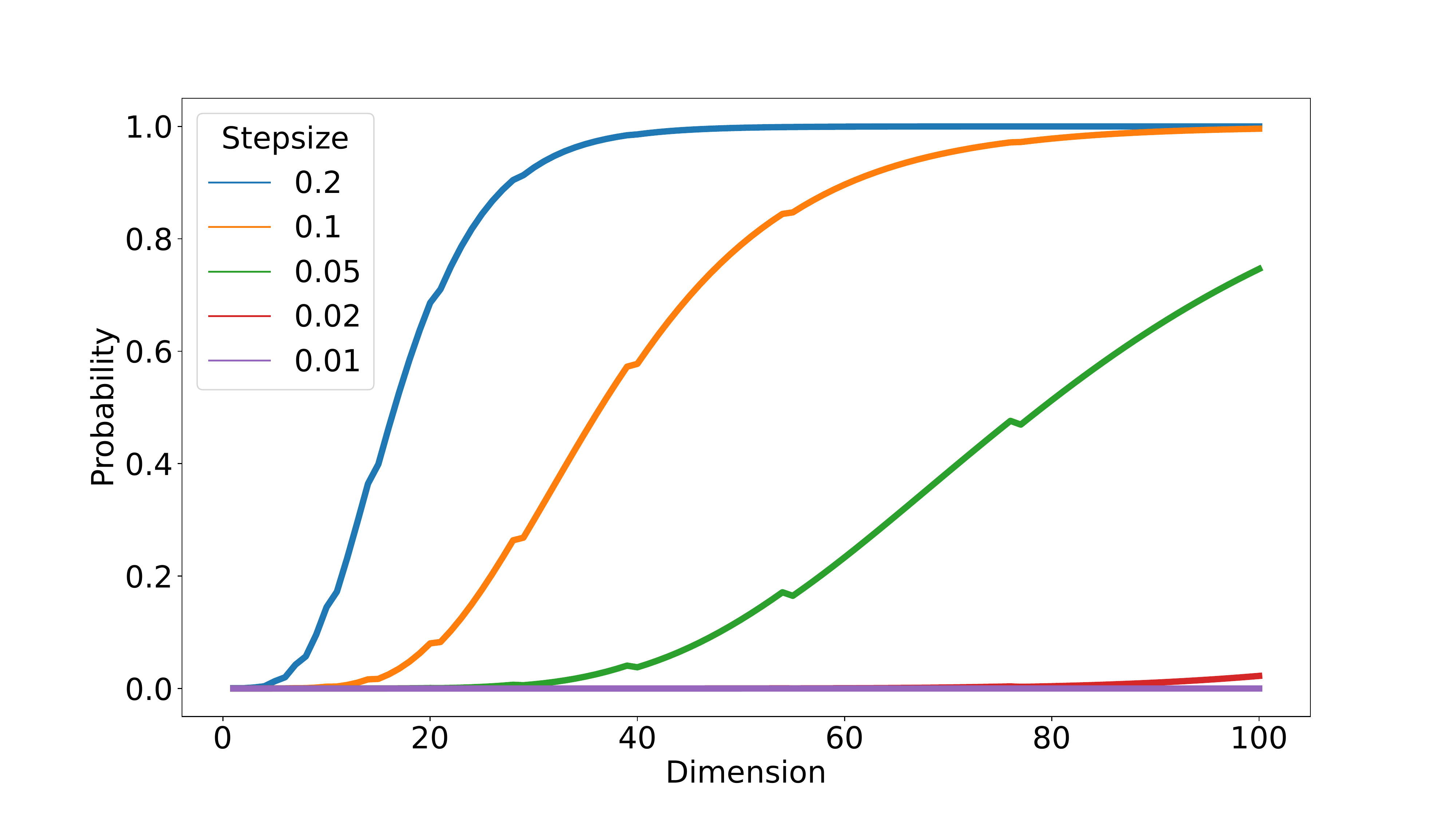}
    \caption{Probability of the whole initial CMA-ES population of size $4 + 3 \log_{10}(d)$ to be infeasible if the center of mass is initialized uniformly at random, for increasing dimensionality $d$ and various values of stepsize $\sigma$.} 
    \label{fig:prob_inf}
\end{figure}

This increase in bound-violations happens due to the higher probability of having a full population of infeasible solutions directly in the first iteration since having a center of mass close to the domain boundary would more likely result in infeasible points when a population is generated from the sampling distribution. Such probability can be estimated as follows:
\begin{enumerate}
    \item 
    In one dimension, the probability of generating a point inside the bounds can be calculated as follows (assuming a normalization of the domain):
        $P=\Pr(0 \leq X \leq 1 \mid X \sim \mathcal{N}(\mu, \sigma^2)) = F_{X}(1) - F_{X}(0) = \frac{1}{2}\left( \text{erf}\left(\frac{1 -\mu}{\sigma\sqrt{2}}\right) + \text{erf}\left(\frac{\mu}{\sigma\sqrt{2}}\right)\right)$, where $F_X(x)$ is the CDF at $x$ and $\text{erf}$ is the error function.
        
    \item     To get the probability for $\mu\sim\mathcal{U}(0,1)$, we take $C=\int_0^1 P \,d\mu $. 
    Since each dimension is identical, and we only need to be infeasible in one to get an infeasible solution, we then get the probability of a single point generated by this distribution to be infeasible to be $1-C^d$ where $d$ is the dimension. 
    \item Then, since we have a default population size of $4 + 3 \log_{10}(d)$, we get the total probability of being infeasible as seen in Figure~\ref{fig:prob_inf}.
\end{enumerate}

From the results in Figure~\ref{fig:prob_inf}, we can see that the probability for the whole initial population to be infeasible grows quickly as the dimensionality increases. To compensate for this, the stepsize has to be reduced significantly. This results in CMA-ES being a very local optimization algorithm, indicating a need to augment the algorithm with constraint-handling strategies and/or mechanisms facilitating exploration. 

\subsection{Stepsize and infeasibility for strategies of dealing with infeasible solutions}\label{secondEx}

\begin{figure*}
    \centering
    \includegraphics[height=0.255\textwidth,trim=9mm 21mm 6mm 5mm,clip]{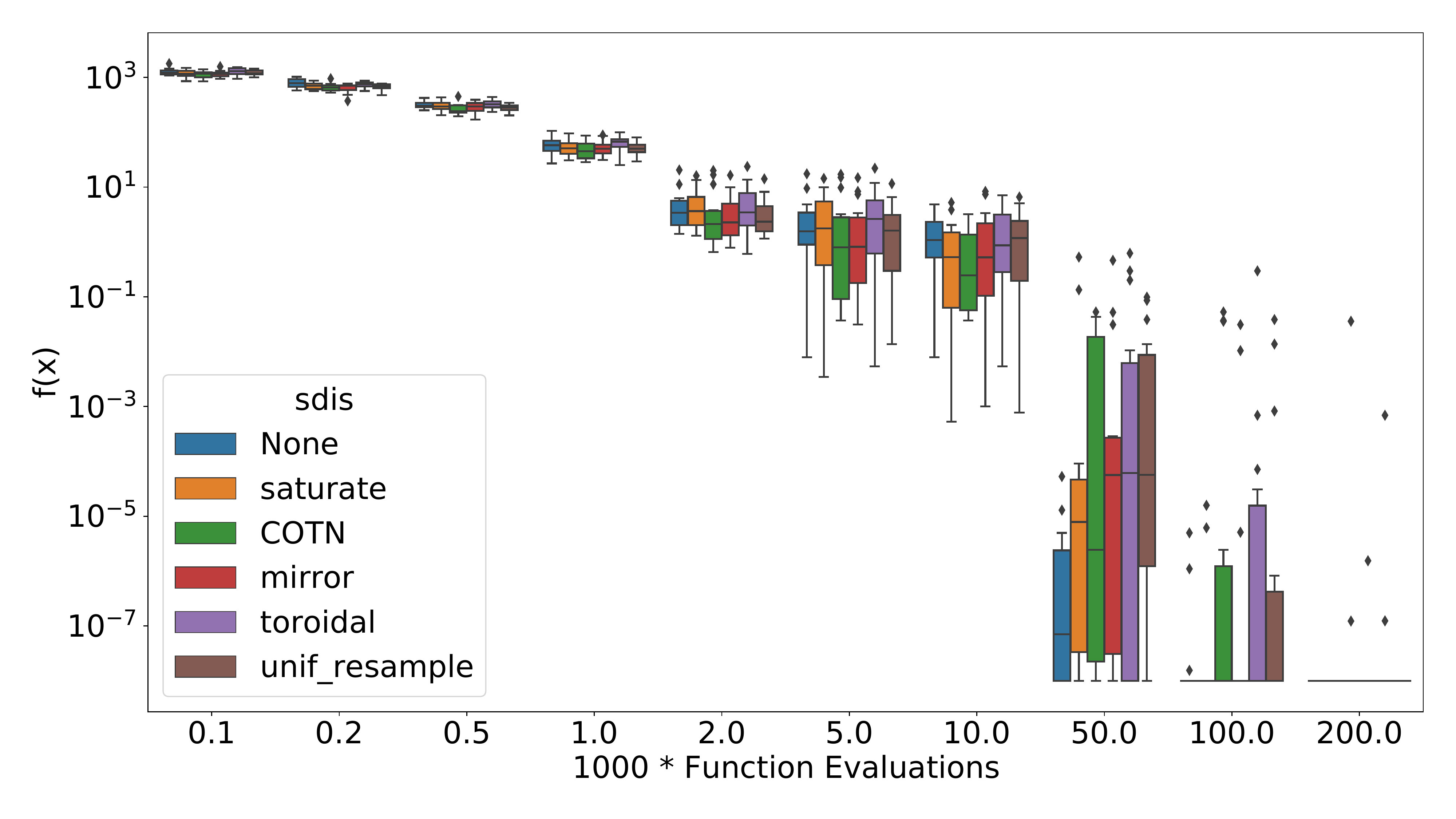}
    \includegraphics[height=0.255\textwidth,trim=17mm 21mm 6mm 5mm,clip]{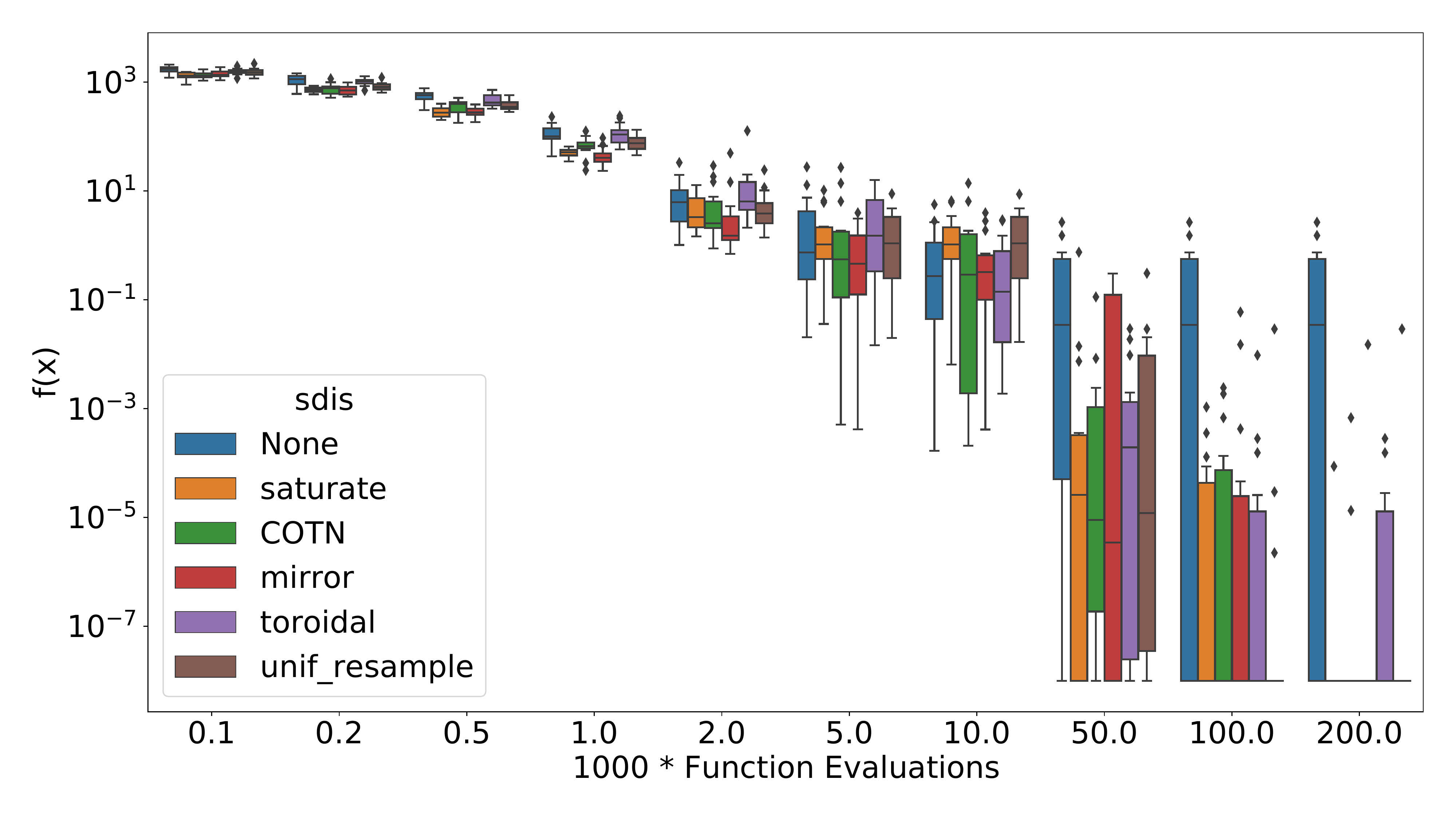}
    \caption{Performance of CMA-ES with center initialization on F13 of BBOB (left) and SBOX-COST (right) in 20D at various budgets (x-axis in thousands of function evaluations).} \label{fig:perf_13_20}
\end{figure*}

\begin{figure*}
    \centering
    \includegraphics[height=0.255\textwidth,trim=9mm 21mm 6mm 5mm,clip]{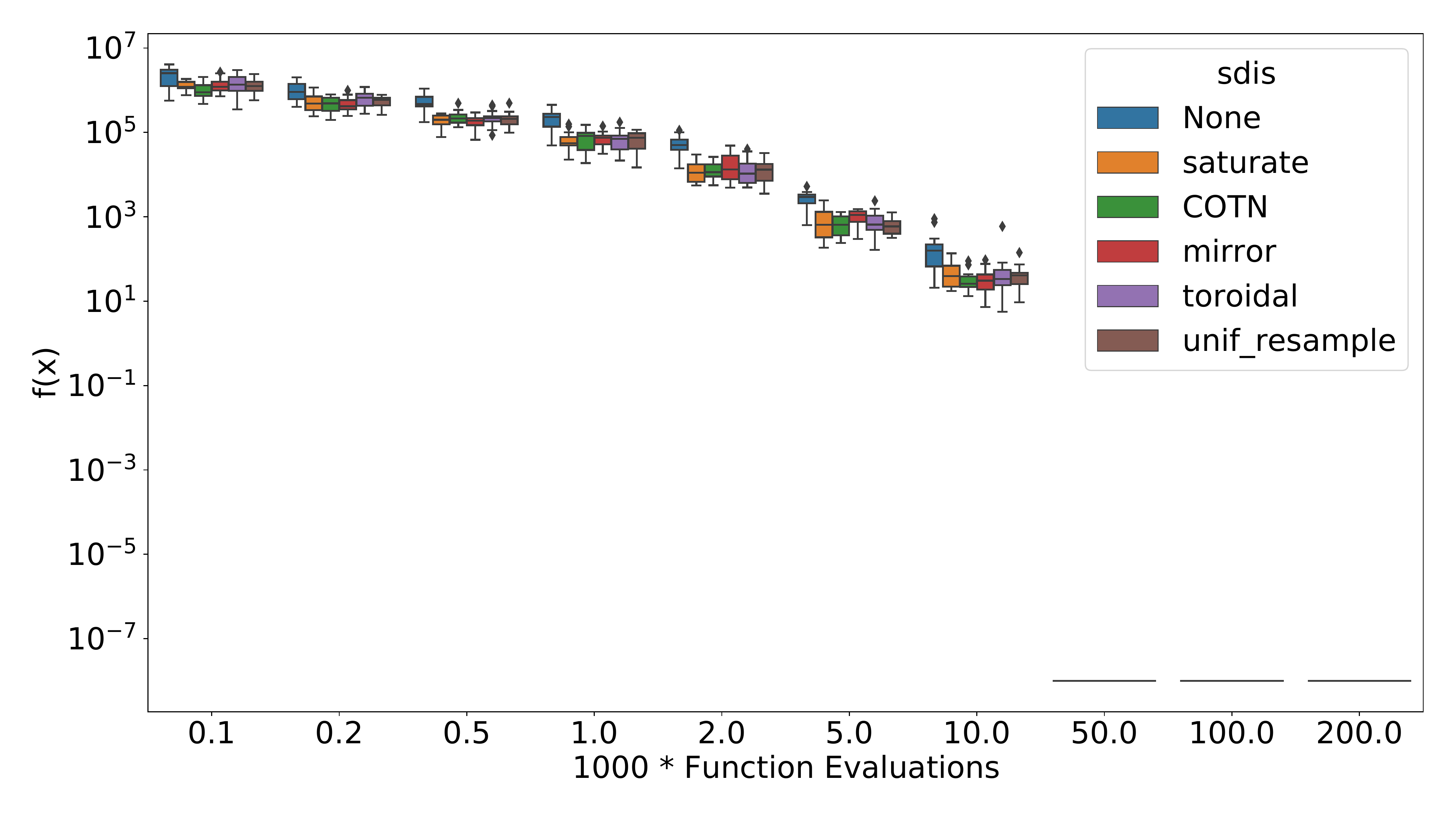}
    \includegraphics[height=0.255\textwidth,trim=17mm 21mm 6mm 5mm,clip]{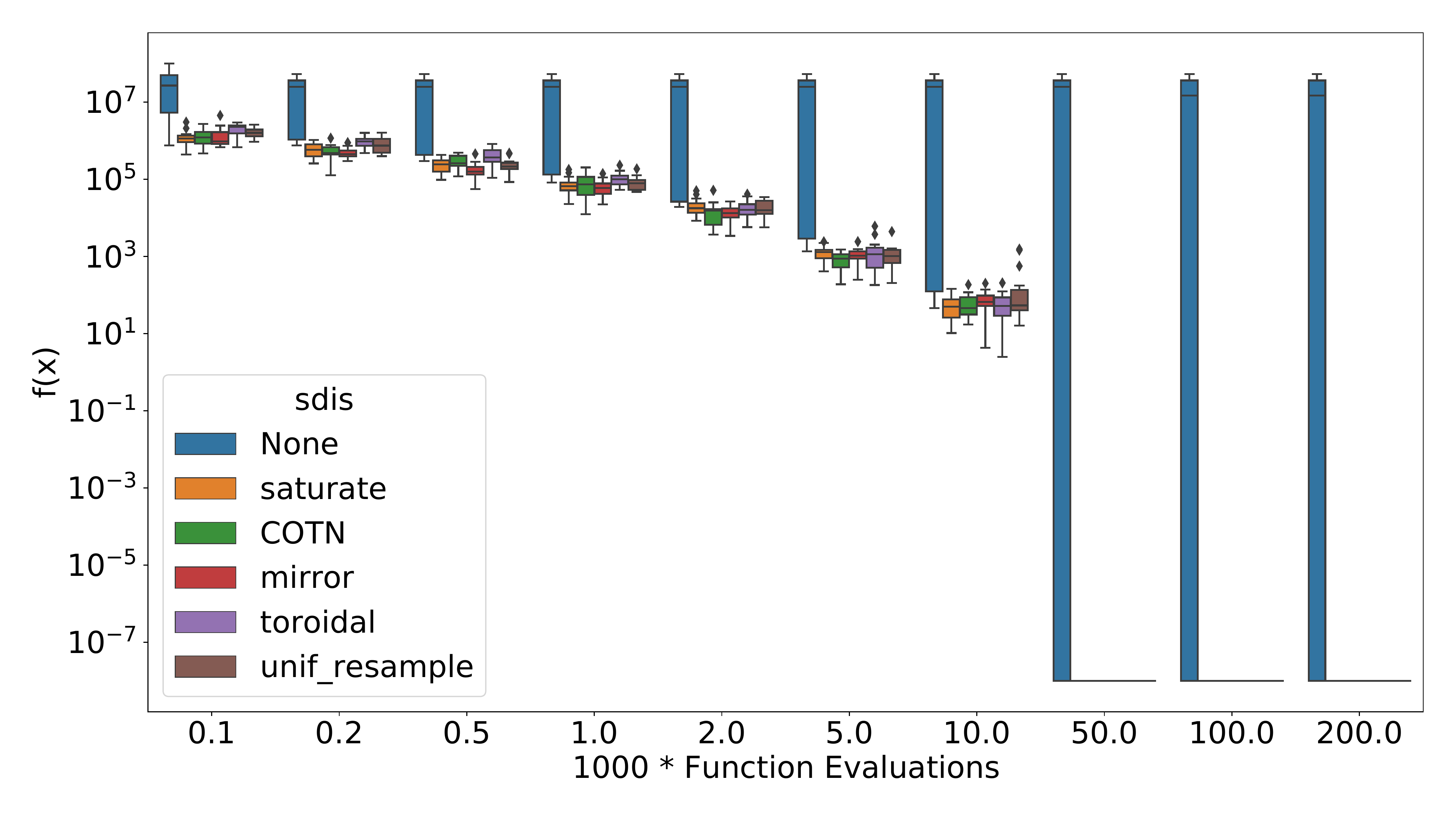}
    \caption{Performance of CMA-ES with uniform initialization on F2 of BBOB (left) and SBOX-COST (right) in 20D at various budgets (x-axis in thousands of function evaluations). } \label{fig:perf_2_20} 
\end{figure*}

\begin{figure}
    \centering
    \includegraphics[height=0.255\textwidth,trim=9mm 21mm 9mm 5mm,clip]{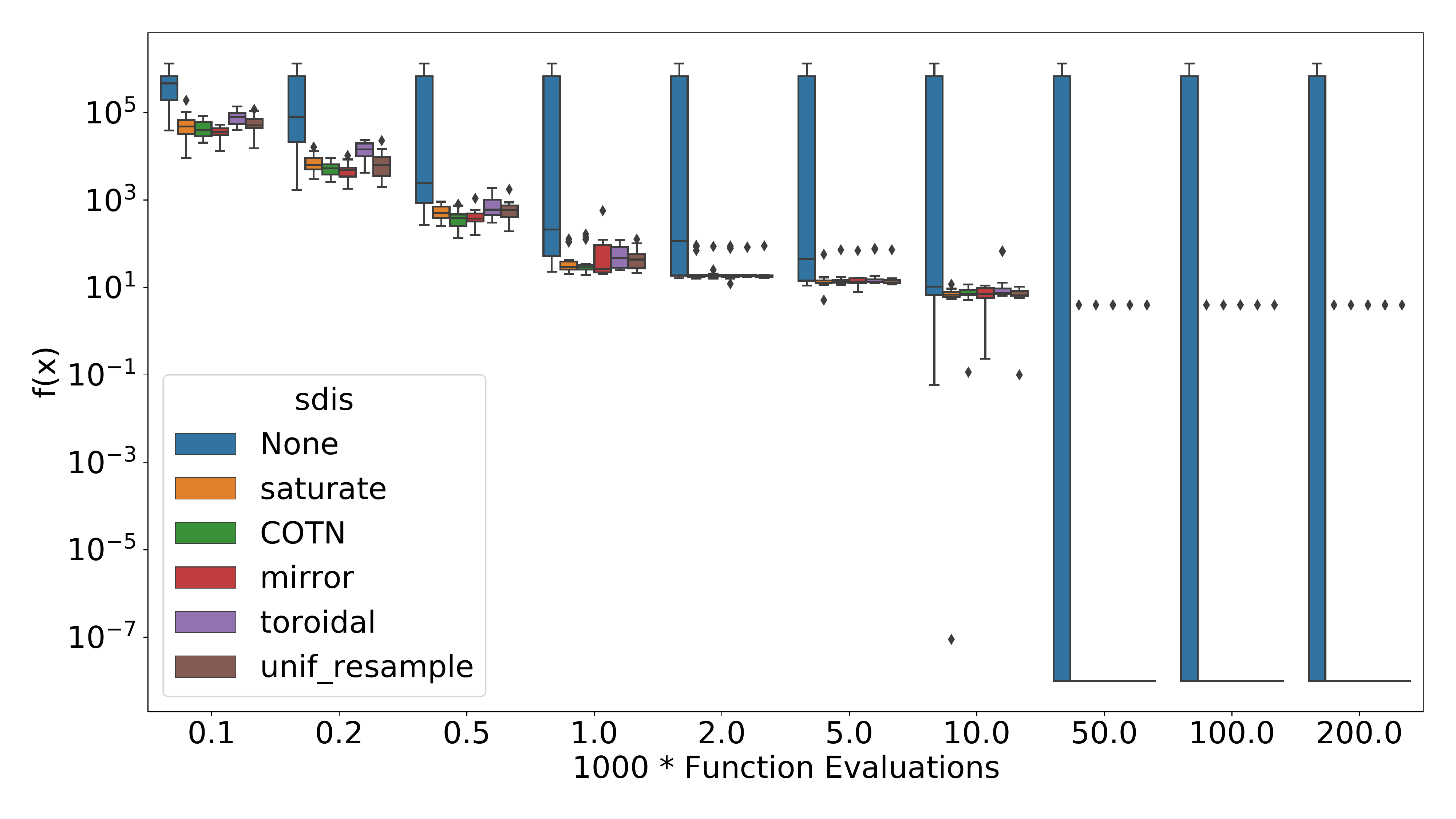}
    \caption{Performance of CMA-ES with uniform initialization on F8 of SBOX-COST in 20D at various budgets (x-axis in thousands of function evaluations).} \label{fig:perf_8_20} 
\end{figure}

\begin{figure}
    \centering
    \includegraphics[width=0.49\textwidth,trim=9mm 20mm 4mm 5mm,clip]{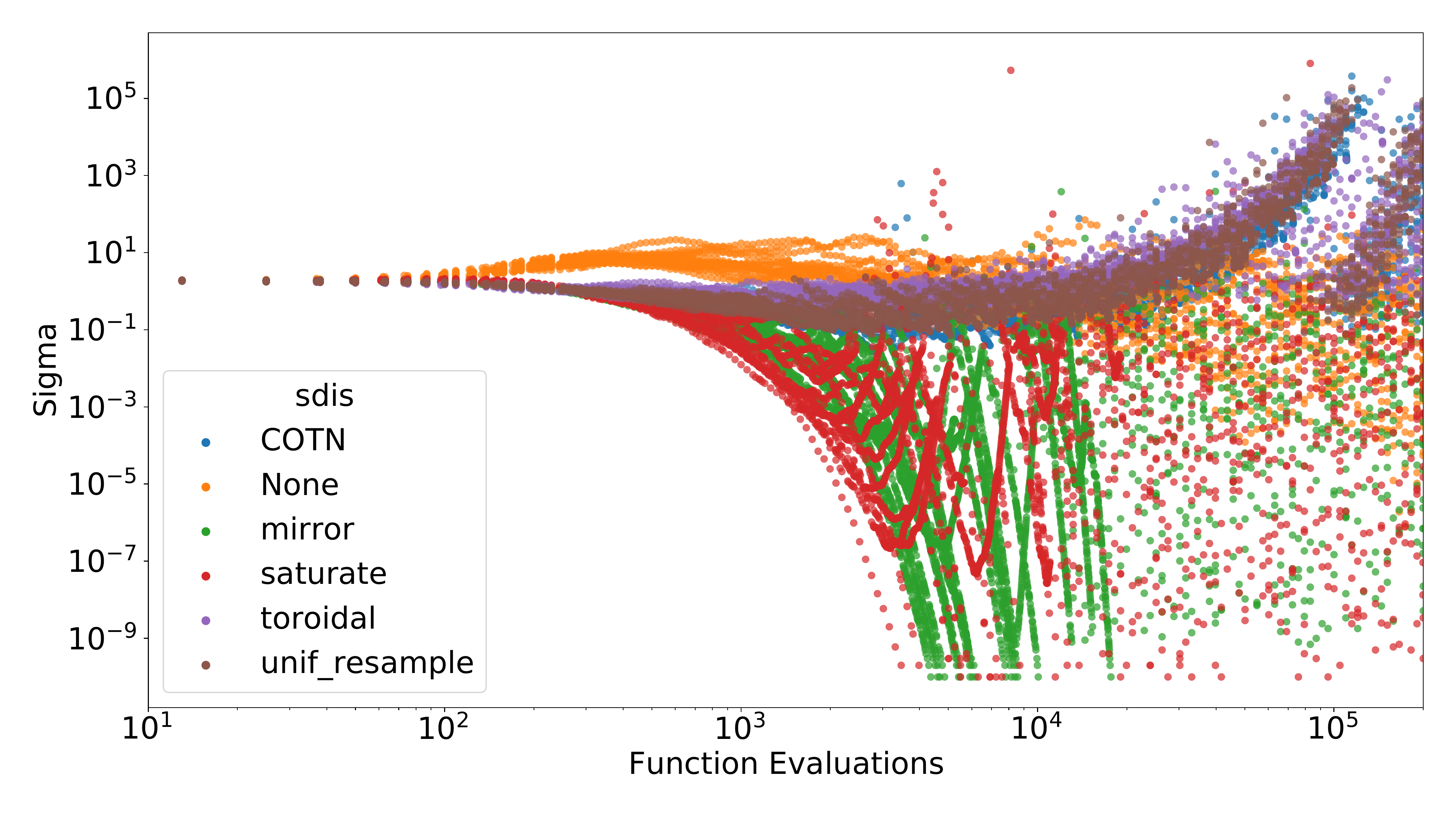}
    \includegraphics[width=0.49\textwidth,trim=9mm 11mm 4mm 5mm,clip]{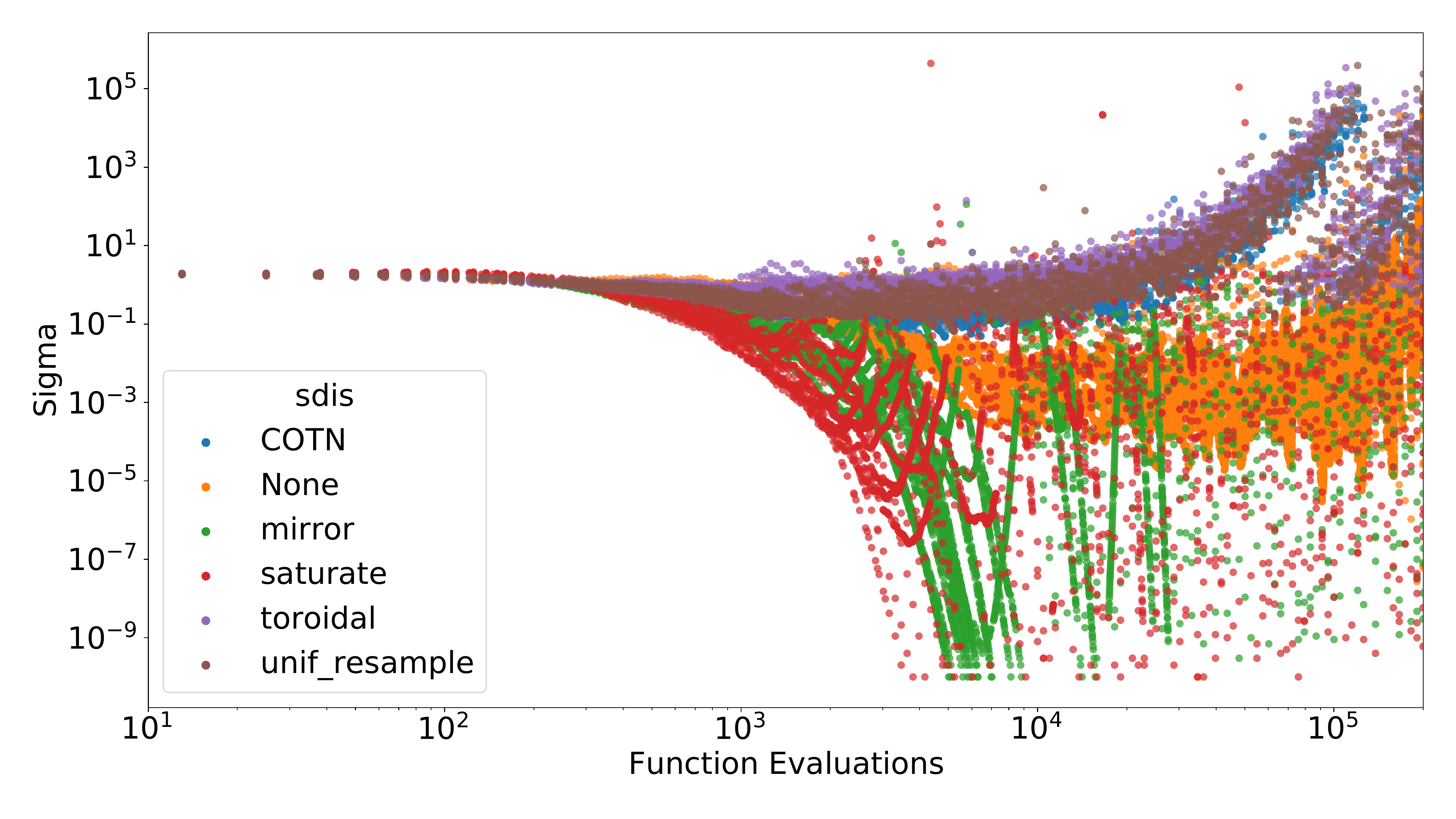}
    \caption{Evolution of stepsize of center-initialized CMA-ES with different strategies of boundary handling on F5 of BBOB (top) and SBOX-COST (bottom) in 20D.}
    \label{fig:sigma_evolv}
\end{figure}

In this experiment, we consider additional boundary handling strategies and analyze in more detail how they affect the stepsize and the number of infeasible solutions. Figure~\ref{fig:perf_13_20} shows the effect of enforcing box-constraints for various strategies on F13 with center initialization: On BBOB (left), there are few differences between the strategies, with \texttt{None} perhaps having a slight advantage; however, on SBOX-COST, the \texttt{None} strategy performs clearly worse. When the initialization is random, this difference becomes more evident (see Figures~\ref{fig:perf_2_20} and~\ref{fig:perf_8_20}), for the reasons discussed earlier. While the differences in performance between different SDIS are minor, Figures~\ref{fig:perf_13_20},~\ref{fig:perf_2_20} and~\ref{fig:perf_8_20} seem to suggest that the more disruptive mechanisms (uniform resampling and toroidal) generally perform worse than the alternatives. 

To gain a better understanding of the worst-case impact of boun\-dary constraints on the performance of CMA-ES, we zoom in on the linear slope (F5), which we observed in Figure~\ref{fig:res_ert} to be impacted significantly when changing from BBOB to SBOX. Additionally, we plot in Figure~\ref{fig:sigma_evolv} the evolution of the stepsize ($\sigma$) for the CMA-ES with center-initialisation and all variants of SDIS studied here. By comparing the figures between BBOB and SBOX, we clearly see that only the \texttt{None} SDIS shows differences in behaviour, since for all other mechanisms points outside the boundaries are never explicitly evaluated. For this \texttt{None} strategy clear differences can be observed: for the SBOX-version of F5, stepsize initially decreases, and keeps varying widely from there. For the BBOB version however, stepsize quickly grows from the beginning of the search. This can be explained by considering that increasing the step-size is very beneficial, as even points generated outside the bound will usually be improvements, as long as the direction is following the slope. So, the larger the steps, the more improvements are made. Then, as soon as a step outside the bound in the right corner is made, the problem is solved since any point beyond there has better fitness than the global optimum.

In addition to the differences observed for the \texttt{None} strategy, Figure~\ref{fig:sigma_evolv} further highlights the differences between the other SDIS. In particular, the more disruptive mechanisms (\texttt{unif\_resampl} and \texttt{toroidal}) show a continuous increase of stepsize. This can be understood by considering that as soon as a point near the optimum, but outside the bound, is generated, it is placed at a very large distance from the center of mass. If this happens often enough, the update mechanism will use these points in the update, even though they fall completely outside of the assumed normal distribution. This leads to an increase in stepsize, and a very poor performance of the CMA-ES. Also of note are the patterns created by the \texttt{saturate} mechanism. In this case, it starts converging as usual, but when it is close enough to the border, the saturation makes sure that even points generated with large step-sizes are mapped to good points often enough, leading to an increase in population size.

Next, we analyze the ratio of solutions generated that violate the box-constraints with respect to the total. Without a boundary handling strategy, those solutions may give an advantage when not enforcing box-constraints (BBOB) while they waste evaluations in SBOX. Figure~\ref{fig:oob} shows this ratio for center (left) and uniform (right) initialization, clearly showing that the ratio is higher for the latter. We can also observe that the ratio differs greatly between functions, being quite small for F9, F18 and F24. One would expect that without a boundary handling strategy (\texttt{None}), CMA-ES will avoid evaluating solutions outside the bounds. However, the plots show that this is not always the case. In fact, with random initialization, there is a high variance and the actual ratio  likely depends on the initial population of each run. 

\section{Conclusions}\label{conclusion}

In this paper we have benchmarked a variety of configurations of the modular CMA-ES on SBOX-COST, which is a variant of the BBOB suite that enforces box-constraints. 

  Our results show that, with strict box-constraints, the lack of a strategy for handling box-constraints often leads to worse results. This effect is stronger when CMA-ES uses random initialization rather than center initialization, presumably because the former has a higher probability of generating infeasible solutions at the start of the run. Indeed, we observe that a higher ratio of infeasible solutions is generated with random initialization. We also analyzed the evolution of the stepsize within each run and the results show why CMA-ES can exploit the lack of boundary constraints to quickly solve the linear slope problem. 
  Finally, we analyzed the effect of handling box-constraints in the BBOB suite. In some functions of BBOB, CMA-ES successfully uses the information provided by infeasible solutions to guide the search more effectively, thus applying boundary handling sometimes leads to worse performance. In general, our results show clear differences between benchmarking algorithms with (SBOX) and without (BBOB) enforcing box-constraints, thus leading to different conclusions about the performance of various CMA-ES variants.

  Results presented in this paper confirm the need for the introduction of a feasibility-enforcing operator within a wider class of heuristic optimisers as pointed out in~\cite{Kononova2022importance}. However, in the case of CMA-ES, enforcing feasibility after the application of generating operator appears to be suboptimal since a sampling operator can potentially significantly increase the ratio of infeasible solutions generated throughout the search. Therefore, as a point of future research, we see the potential for designing new adaptive strategies of dealing with infeasible solutions that truncate the sampling distribution depending on the current location of its mean within the box-constrained domain. Such strategy can be applied either to only infeasible dimensions as done in most cases considered in this paper or to all dimensions of an infeasible solution to preserve the search direction~\cite{Kononova2022importance} prescribed based on the points successfully sampled so far; see~\cite{kreischer2017} for a similar idea for the Differential Evolution algorithm.

\subsubsection*{Reproducibility}
Source code and the data generated in this study are available from \url{https://doi.org/10.5281/zenodo.7649077}.

\bibliographystyle{ACM-Reference-Format}
\bibliography{references}

\end{document}